\pdfoutput=1

\documentclass[11pt]{article}

\usepackage{EMNLP2023}

\usepackage{times}
\usepackage{latexsym}

\usepackage[T1]{fontenc}

\usepackage[utf8]{inputenc}

\usepackage{microtype}

\usepackage{inconsolata}

\usepackage{url}

\usepackage{enumitem}
\usepackage{booktabs}

\usepackage{graphicx}
\usepackage{colortbl}

\usepackage{wrapfig}
\usepackage{tipa}

\usepackage{amsmath}

\newcommand\blfootnote[1]{%
  \begingroup
  \renewcommand\thefootnote{}\footnote{#1}%
  \addtocounter{footnote}{-1}%
  \endgroup
}

\title{CCGen: Explainable Complementary Concept Generation in E-Commerce}

\author{Jie Huang$^{1*}$ $\quad$ Yifan Gao$^{2\dag}$ $\quad$ Zheng Li$^2$ $\quad$ Jingfeng Yang$^2$ $\quad$ Yangqiu Song$^3$ $\quad$ Chao Zhang$^4$ \\ \textbf{Zining Zhu$^5$ $\quad$ Haoming Jiang$^2$ $\quad$  Kevin Chen-Chuan Chang$^1$ $\quad$ Bing Yin$^2$} \\
 $^1$University of Illinois at Urbana-Champaign $\quad$ $^2$Amazon \\
 $^3$Hong Kong University of Science and Technology \\ $^4$Georgia Institute of Technology $\quad$
 $^5$University of Toronto \\
 \texttt{\{jeffhj,kcchang\}@illinois.edu}, \texttt{\{yifangao,amzzhe,jingfe,jhaoming,alexbyin\}@amazon.com} \\ 
\texttt{yqsong@cse.ust.hk}, \texttt{chaozhang@gatech.edu}, \texttt{zining@cs.toronto.edu}}

\begin{document}
\maketitle
\begin{abstract}
We propose and study \textbf{C}omplementary \textbf{C}oncept \textbf{Gen}eration (\textbf{CCGen}): given a concept of interest, e.g., ``\textit{Digital Cameras}'', generating a list of complementary concepts, e.g., \textit{1) Camera Lenses 2) Batteries 3) Camera Cases 4) Memory Cards 5) Battery Chargers}. CCGen is beneficial for various applications like query suggestion and item recommendation, especially in the e-commerce domain. To solve CCGen, we propose to train language models to generate ranked lists of concepts with a two-step training strategy. We also teach the models to generate explanations by incorporating explanations distilled from large teacher models. Extensive experiments and analysis demonstrate that our model can generate high-quality concepts complementary to the input concept while producing explanations to justify the predictions.
\blfootnote{$^*$Work done during internship at Amazon. $^\dag$To whom correspondence should be addressed.}
\end{abstract}

\maketitle

\section{Introduction}

Concepts play a crucial role in our daily lives, as they help us comprehend and interpret the world around us. 
In the e-commerce domain, for example, concepts are prevalent in both customer queries and the products being offered. 
When a user wants to take photos, he/she may think of the concept of ``\textit{cameras}'' and use this concept to form search queries to find products related to this concept. 

When users explore concepts, they are not only interested in similar concepts, but also in complementary ones. 
For example, a user who is interested in ``\textit{digital cameras}'' may also want to explore complementary concepts such as ``\textit{memory cards}'', ``\textit{camera cases}'', or ``\textit{complete tripods}'', more than similar ones like ``\textit{film cameras}'' to fulfill his/her overall intent.
However, complementary concepts may not be immediately obvious to the user, 
as they are based on functionality and complementarity.

\begin{figure}[tp!]
    \centering
    \includegraphics[width=\linewidth]{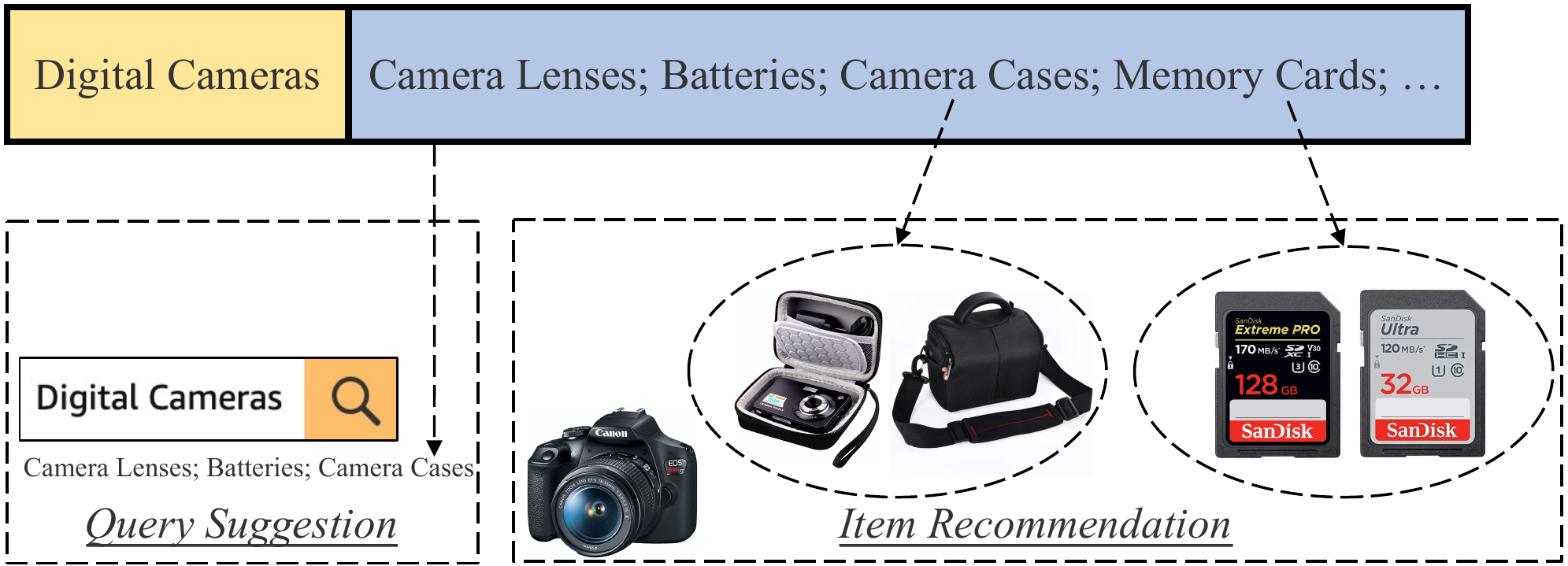}
    \caption{A demonstration of Complementary Concept Generation and its applications. Here ``\textit{Digital Cameras}'' is the input and ``\textit{Camera Lenses; ...}'' is the output.} 
    \label{fig:GCP}
    \vspace{-5mm}
\end{figure}

Inspired by this, we propose a new task named \textbf{C}omplementary \textbf{C}oncept \textbf{Gen}eration (\textbf{CCGen}) -- generating concepts that are complementary to an input concept. 
For example, when given a concept ``\textit{digital cameras}'', the expected output would be concepts such as ``\textit{camera lenses}'', ``\textit{camera cases}'', and ``\textit{memory cards}''. 
CCGen has great potential to facilitate many important applications, including query suggestion~\citep{cao2008context,ma2010diversifying,ooi2015survey}, item recommendation~\citep{hao2020p,yan2022personalized}, and exploratory search~\citep{marchionini2006exploratory}.
This is particularly useful in the e-commerce domain, as it helps users to discover new or previously unknown concepts, navigate between related concepts, and combine concepts to serve the overall intent (refer to Figure~\ref{fig:GCP} for some examples).

Although research on entity/concept relationships has been ongoing for a significant period, to our knowledge, there is currently no work specifically addresses our problem. For example, CCGen can be connected to finding related concepts, but most existing research in this area focuses on identifying semantically relevant concepts in general, ``\textit{dog}'' is related to ``\textit{cat}''~\citep{mikolov2013distributed,yamada2020wikipedia2vec,wang2007language}. 
To connect with the study of a specific type of relation like ``\textit{is a}'' relation, our problem can be viewed as studying the ``\textit{is complementary to}'' relation. However, while there have been many studies on discovering hypernymy~\citep{hearst1992automatic,snow2004learning,roller2018hearst} and constructing taxonomies~\citep{miller1995wordnet,liu2012automatic} for ``\textit{is a}'' relation, there is currently no research specifically on ``\textit{is complementary to}'' relation.

There are several challenges for CCGen.
\textbf{C1}: First, generating complementary concepts requires relational and commonsense knowledge about the concepts, which cannot be simply achieved by measuring concept similarity. 
For instance, ``\textit{film cameras}'' is similar to ``\textit{digital cameras}'', but not as complementary to ``\textit{digital cameras}'' as ``\textit{memory cards}''. 
\textbf{C2}: Second, acquiring large amounts of data for CCGen can be difficult, unlike traditional query suggestion, which can rely on readily available click-through and session data~\citep{cao2008context,ooi2015survey}.
\textbf{C3}: Third, the relationships between complementary concepts to the input concept may not be intuitive, so that users may not understand why these concepts are complementary.

To overcome these challenges, we propose to generate complementary concepts and simultaneously explain why each concept is complementary to the given one.
For \textbf{C1}, we leverage the rich relational and commonsense knowledge stored in the parameters of pre-trained language models (PLMs) \citep{petroni2019language,roberts2020much,zhou2020evaluating,jiang2020can,jiang2020x,huang2022open} and solve the task in a novel form of list generation.
Specifically, we fine-tune PLMs to generate ranked lists of concepts, e.g., ``\texttt{1) Camera Lenses 2) Batteries 3)
Camera Cases 4) ...}'' with ``\textit{Digital Cameras}'' as the input.
To overcome \textbf{C2}, we manually craft a concept set of size $\sim$$7k$ and build a dataset upon it by leveraging existing user behavior data from \citet{ni2019justifying}. 
To mitigate overfitting caused by limited training data, we introduce an effective two-step training strategy with unordered and ordered list generation objectives, meaning, training on unordered lists first and then on ordered lists.
To make the predictions more explainable (\textbf{C3}), we attempt to teach the models to generate explanations to justify the predictions while making the predictions by incorporating free-text explanations distilled from large language models, e.g., GPT-3~\citep{NEURIPS2020_1457c0d6} or Meta OPT~\citep{zhang2022opt}.

To verify the effectiveness of the proposed method,
we introduce two metrics to measure the quality of generated concepts and conduct extensive evaluation and analysis.
From our experiments, we find that 1) generative language models have a great ability to generate complementary concepts in our proposed list generation form;
2) the proposed two-step training strategy can greatly improve model performance;
3) by incorporating explanations distilled from teacher models, the models can make better predictions while generating explanations to justify the predictions.
In summary, the novel CCGen task, the methods proposed, and the findings presented have significant potential to offer new perspectives for future research and facilitate related downstream applications.

\section{Complementary Concept Generation}

The goal of Complementary Concept Generation (CCGen) is to generate concepts that are complementary to a given concept.
Formally, given a concept $x$, the expected output is a list of concepts $\mathcal{R}_x = [y_1, y_2, \dots, y_k]$.
In this paper, we focus on complementary concept generation in the e-commerce domain. E.g., given an e-commerce concept ``\textit{Digital Cameras}'', the output is a concept list: [``\textit{Camera Lenses}'', ``\textit{Batteries}'', ``\textit{Camera Cases}'', $\dots$].
Since not all concepts are ``equally'' relevant and complementary to the given one,
we consider the order of the list by ranking the concepts based on their ``frequency of co-purchase'' with the given concept in the user behavior data (more details are in \S\ref{sec:data}).

\begin{figure*}[tp!]
  \centering
\includegraphics[width=0.97\linewidth]{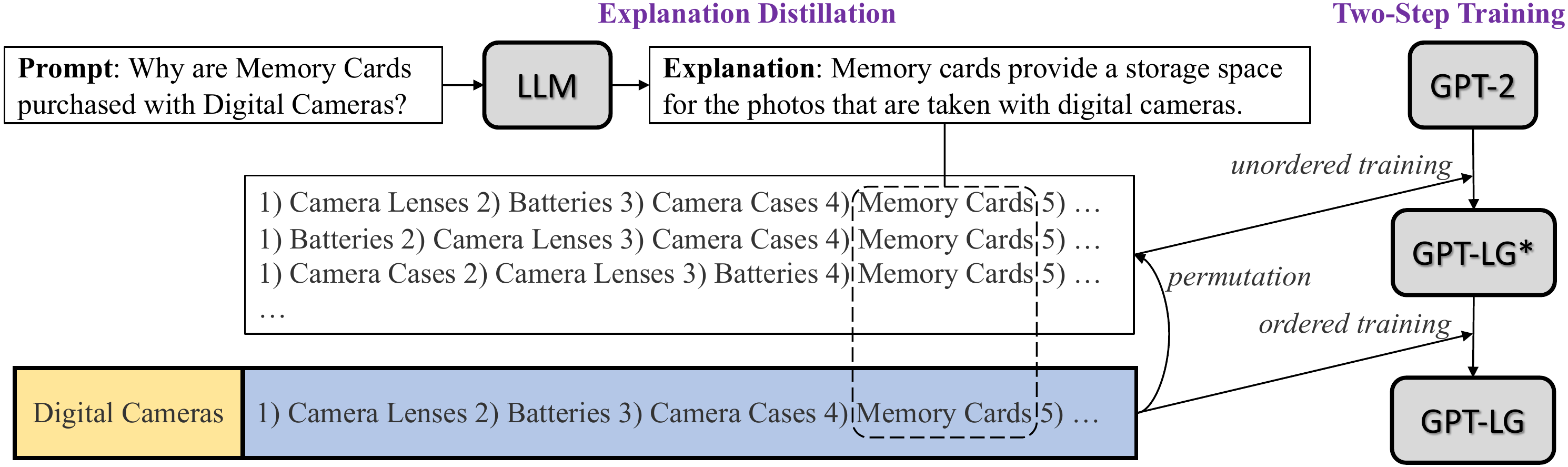}
    \caption{The overview of the proposed method. GPT-LG is trained with two steps: 1) \textit{ unordered training}, where the model is trained to produce different permutations of the target ranked list; 2) \textit{ordered training}, where the model is trained to rank the concepts. By leveraging explanations distilled from large language models (LLMs), GPT-LG can also generate free-text explanations to justify its predictions while generating the concepts.} 
    \label{fig:method}
    \vspace{-2mm}
\end{figure*}

To solve CCGen, we propose to train language models to generate ranked lists of concepts 
and free-text explanations to justify the predictions simultaneously. The overview of our method is illustrated in Figure~\ref{fig:method}.

\subsection{Complementary Concept Generation via List Generation}
\label{sec:lg}

When we discover complementary concepts, we rely on our knowledge.
Similarly, to do CCGen,
the model needs relational and commonsense knowledge about the concepts, 
e.g., $e_0$: ``\emph{Memory cards}'' are purchased with ``\emph{digital cameras}'' because they provide a storage space for the photos that are taken with the camera.

Recent studies have demonstrated that pre-trained language models contain a significant amount of relational and commonsense knowledge \citep{petroni2019language,roberts2020much,zhou2020evaluating,jiang2020can,jiang2020x,huang2022open}, and such knowledge
is critical for CCGen.
For instance, $e_0$ is actually generated by GPT-3 with only prompt ``\texttt{Why are Memory Cards purchased with Digital Cameras?}''. This indicates these models even have the ability to explain their predictions explicitly.

To leverage rich relational and commonsense knowledge in pre-trained language models directly, we propose to solve the CCGen problem with the objective of language modeling directly. To achieve this, we convert a list $\mathcal{R}_x$ to a text sequence $s =$ ``1) $y_1$ 2) $y_2$ ... $k$) $y_k$'' (we set $k=5$ by default). 
And then, the probability of the output can be computed by language models auto-regressively:
\begin{equation}
\label{eq:lm}
P_\theta(s|x) = \prod_{i=1}^{|s|} P_\theta(s_i|s_{0:i-1}, x),
\end{equation}
where $|s|$ is the length of $s$, $s_i$ is the $i$th token of $s$, and $s_0$ is a special start token. 
Intuitively, the ranking of suggested concept can be encoded as the serial number, i.e., ``$i)$'', in the sequence.
Particularly, we employ GPT-2 \citep{radford2019language}, a transformer-based causal language model pre-trained on large corpora. We enable GPT-2 for CCGen by fine-tuning the model to produce ranked lists of concepts.
Specifically, we apply the following scheme: 
``\texttt{[SOS] $prompt(x)$ 1) $y_1$ 2) $y_2$ ... $k$) $y_k$ [EOS]}''
\footnote{We use prompt ``$x$ are purchased with'' by default and add ``[SOS]'' and ``[EOS]'' to indicate the start and end of the sequence. In our experiments, we find that the effect of prompt selection (e.g., replacing ``purchased'' with ``bought'' or ``co-purchased'') is very small.},
e.g., ``\texttt{[SOS] Digital Cameras are purchased with 1) Camera Lenses 2) Batteries 3) Camera Cases 4) Memory Cards 5) Battery Chargers [EOS]}''.
After fine-tuning on the training data, the model learns to generate a ranked list of concepts with a given concept. Formally, the formula in Eq.~\eqref{eq:lm} is modified as
\begin{equation}
P_\theta(s|x) = \prod_{i=1}^{|s|} P_\theta(s_i|s_{0:i-1}, prompt(x)).
\end{equation}
We name the above model as \textbf{GPT-LG}, where ``GPT'' refers to ``\textbf{GPT}-2'' and ``LG'' denotes ``\textbf{L}ist \textbf{G}eneration''.

\subsection{Two-Step Training} 

Based on the above design, we can enable a language model to generate a ranked list of concepts that are complementary to the given one. However, we find that there still exist two limitations: First, the model tends to \textit{overfit} if the number of training examples is limited. Second, the model is not \textit{robust} since the next prediction largely relies on the previous predictions. If the previous concept, e.g., the second position, is predicted wrongly, the next concept, e.g., the third position, will tend to be mispredicted.

Therefore, we propose to train the model in two steps: \textit{unordered training (ut)} and \textit{ordered training (ot)}.
In the unordered training step, we train the model to produce target concepts without considering the order of the concepts. Specifically, we shuffle the order of the ranked list to generate different permutations, e.g., ``1) $y_2$ 2) $y_k$ ... $k$) $y_1$'', and fine-tune PLMs to produce these permutations. In this way, we can acquire much more training examples, so the training is less prone to overfitting. Besides, shuffling breaks up the patterns of ranked lists; therefore, the trained model will predict the next concept more based on the input concept than the previous concepts. 
Formally, the objective of the unordered training step is:
\begin{equation}
\label{eq:ut}
\max_\theta \quad \sum_{x} \sum_{z \in \mathcal{Z}} \log P_\theta(z|x),
\end{equation}
where $\mathcal{Z}$ is the set of sampled permutations for $s$.
After the unordered training, we perform the ordered training with the training examples in \S\ref{sec:lg}. 
With this step, the model learns to rank the concepts in its generation.

\subsection{Explanation Distillation}

Besides generating concepts, we try to enable models to explain their predictions, e.g., by providing free-text explanations.
There are two main benefits to making CCGen more explainable. 
First, explanations can help users understand why the generated concept is complementary to the present one, which is not always obvious.
Second, explanations can validate the predictions. We do not expect the model to generate something that it cannot explain. This can also potentially benefit the performance as suggested by \citet{lampinen2022can}, since the model tends to predict concepts that it can explain.

However, annotating explanations is difficult and it requires expensive human efforts \citep{jang2021training}. 
Recent studies show that large language models (LLMs) such as GPT-3 can generate high-quality free-text rationales or explanations \cite{wiegreffe2021reframing,huang2022reasoning}. 
Based on this, we propose to distill explanations from LLMs, namely \textit{teacher models}, and incorporate them into the base model \citep{li2022explanations,shridhar2022distilling,magister2022@teaching}. 
Specifically, we fine-tune the base model to solve a list generation task with the following encoding scheme: ``\texttt{[SOS] $prompt(x)$ 1) $y_1$: $e_1$ 2) $y_2$: $e_2$ ... $k$) $y_k$: $e_k$ [EOS]}'', where $e_i$ is a free-text explanation that explains why $y_i$ is co-purchased with $x$. To acquire the explanations, we query large language models such as OpenAI GPT-3 \citep{NEURIPS2020_1457c0d6} or Meta OPT \citep{zhang2022opt} with prompt ``\texttt{Explain why one product is purchased with the other product.$\backslash$n$\backslash$n Q: Why are $y$ purchased with $x$?$\backslash$n A:}''.
For instance, we prompt OPT
and form the following sequence for \textit{Digital Cameras}:
``\texttt{[SOS] Digital Cameras are purchased with 1) Camera Lenses: The camera lens is the part of the camera that focuses the light from the object into a picture. 2) ... 5) Battery Chargers: Battery Chargers ... are needed to recharge the batteries that are used in Digital Cameras. [EOS]}''
In this way, we enable the base model to provide explanations when generating concept lists.

\section{Experiments}

In this section, we first describe the process of data collection and introduce two evaluation metrics. We then test the performance of the models with different settings and various ablation studies. 

\subsection{ECCGen Dataset}
\label{sec:data}
Since there is no public dataset for complementary concept generation, we build a benchmark dataset, named \textbf{ECCGen} (\textbf{E}-commerce \textbf{C}omplementary \textbf{C}oncept \textbf{Gen}eration), 
based on purchase history records in the public Amazon Review Data \citep{ni2019justifying}.
The idea is that products purchased at the same time are likely to be complements~\citep{mcauley2015inferring,hao2020p,yan2022personalized} and a concept can be associated with many products; thus based on statistics, the two most frequently co-purchased concepts are highly likely to be complementary concepts.

Based on this idea, we manually craft a concept set of size $\sim$$7k$ and build a dataset upon it by leveraging user behavior data from \citet{ni2019justifying}.
We summarize the statistics of the data in Table~\ref{table:data} and refer the reader to Appendix~\ref{sec:data_detail} for more details.

To verify the quality of the dataset, we sample 100 examples from the test set and let three human annotators judge whether the concepts in each list are complementary to the input concept. The results suggest that $\sim$96\% of the concepts are complementary well to the input concept, indicating the overall quality is high.

\begin{table}[tp!]
    \begin{center}
    \begin{tabular}{c|c|ccc}
        \toprule
        & \text{\textbf{\# Concept}} & \textbf{Train} & \textbf{Dev} & \textbf{Test}  \\
        \midrule
        Size & 7,084 & 4,787 & 350 & 702 \\
        \bottomrule
    \end{tabular}
    \end{center}
    \vspace{-2mm}
    \caption{The statistics of ECCGen dataset.}
    \label{table:data}
    \vspace{-4mm}
\end{table}

\subsection{Experimental Setup}

{\flushleft \textbf{Baselines}.} 
Since there is no existing work on complementary concept generation,
we evaluate our methods with different sizes of models and different variants.
We also modify some existing methods and design several baselines as follows:
\begin{itemize}[leftmargin=*,itemsep=1pt]
\item \textbf{GloVe}. To verify that CCGen cannot be solved by concept similarity, we apply compositional GloVe embeddings \citep{pennington2014glove} (element-wise addition of the pre-trained 100d word embeddings) as features of concepts and rank concepts based on cosine similarities.
\item \textbf{KNN}. We apply the K-Nearest Neighbor (KNN) algorithm \citep{fix1989discriminatory} as follows: given concept $x$, we get the $k$ closest concepts in the training data based on GloVe. And then, we collect all the top $5$ complementary concepts of these $k$ closest concepts and rank these concepts by count.
\item \textbf{SVM}. We train an SVM \citep{cortes1995support} binary classifier to rank the concepts. Specifically, we create positive pairs $(x,y)$ where $y$ is in the top $5$ complementary concepts and sample $5$ negative pairs for each positive pair. We concatenate the compositional GloVe embeddings of concepts as the features. Concepts are selected based on confidence of the prediction.
\item \textbf{Item2vec}. Item2vec \citep{barkan2016item2vec,wang2018path} learns item embeddings based on item-item interactions.
We modify Item2vec as follows: we freeze the target embeddings as the compositional GloVe embeddings (to handle concepts that do not appear in the train set) and learn the context embeddings with the same training data of SVM. Concepts are ranked based on the cosine similarities of the target and context embeddings.
\item \textbf{P-Companion}. P-Companion \citep{hao2020p} is a neural-network model for complementary product recommendation. We modify this model to our task by setting $\alpha = 0$, i.e., only complementary type modeling. Similar to Item2vec, we freeze the target embeddings as the compositional GloVe embeddings and train the model with negative sampling. Concepts are selected based on confidence of the prediction.
\item \textbf{GPT-3}. We also compare with GPT-3 (text-davinci-2) \citep{NEURIPS2020_1457c0d6}. Specifically, we randomly sample 5 examples from the train set as the demonstrations and prompt GPT-3 to generate a ranked list for the target concept. Since many concepts generated by GPT-3 are not in the concept set, we map these concepts to the closest ones in the concept set based on GloVe. 
\end{itemize}

{\flushleft \textbf{Metrics}.} We design two metrics for evaluation. 
Since the model may generate duplicate concepts, we calculate accuracy as follows:
\begin{equation}
\resizebox{\hsize}{!}{$ACC@k_m = \quad \frac{1}{|\mathcal{K}|} \sum_{y' \in \mathcal{K}} \mathbf{1}[rank(y'_m) \leq k \land y'_m \notin y'_{1:m-1} ]$,}
\end{equation}
where $\mathcal{K}$ is the test set, $y'_m$ is the $m$-th prediction of the model, $rank(y'_m)$ denotes the rank of $y'_m$ among the all the candidate concepts, and $y'_{1:m-1} = \{y'_1, y'_2, \dots, y'_{m-1}\}$.
We calculate the average as the overall accuracy:
\begin{equation}
ACC@k_{overall_m} = \frac{1}{m} \sum_{i = 1}^m ACC@k_i.
\end{equation}
We also adapt the $nDCG$ metric \citep{jarvelin2002cumulated} to our problem, formulated as:
\begin{equation}
    \resizebox{\hsize}{!}{
    $nDCG_m = \frac{DCG_m}{iDCG_m}, \quad DCG_m = \sum_{i=1}^m \frac{w(y'_i)}{\log_2(i+1)}$,}
\end{equation}
where 
\resizebox{0.73\hsize}{!}{
$
w(y'_i) = 
\begin{cases}
conf(x,y'_i) & \text{if $y'_i \notin y'_{1:i-1}$} \\
0 & \text{otherwise}
\end{cases}
$}
and $iDCG_m = \sum_{i=1}^{m} \frac{w(y_i)}{\log_2(i+1)}$, with $y_i$ as the ground-truth concept at position $i$.
$conf(x,y) = freq(x,y)/freq(x)$, where $freq(x,y)$ is the frequency that $x$ and $y$ are co-purchased and $freq(x)$ is the frequency of $x$.
For these two metrics, $ACC@k$ only cares about whether the predicted concept is a relevant one, and $nDCG$ is sensitive to  the order of concepts.

\begin{table*}[tp!]
\scriptsize
\begin{center}
\begin{tabular}{l|ccccc|c|c|c}
\toprule
Model / Score (\%) & \textbf{1} & \textbf{2} & \textbf{3} & \textbf{4} & \textbf{5} & \textbf{Overall} & \textbf{nDCG} & \textbf{VR} \\
\midrule 
GloVe & 19.37 & 13.82 & 11.54 & 9.69 & 6.55 & 12.19 & 10.12 & - \\
KNN & 31.20 & 26.78 & 22.93 & 18.80 & 17.66 & 23.48 & 21.04 & - \\ 
SVM & 7.83 & 6.55 & 6.13 & 6.41 & 6.98 & 6.78 & 6.77 & - \\ 
Item2vec & 20.51 & 14.25 & 11.11 & 9.97 & 7.12 & 12.59 & 10.97 & - \\
P-Companion & 23.50 & 14.39 & 13.68 & 11.40 & 10.83 & 14.76 & 14.35 & - \\
GPT-3 & 10.97 & 8.69 & 6.41 & 6.98 & 5.98 & 7.81 & 8.01 & - \\ 
\hline
GPT-LG-small & 16.84\tiny{$\pm$0.35} & 7.89\tiny{$\pm$0.48} & 3.76\tiny{$\pm$0.23} & 2.39\tiny{$\pm$0.36} & 1.34\tiny{$\pm$0.07} & 6.44\tiny{$\pm$0.15} & 15.20\tiny{$\pm$0.22} & 86.10\tiny{$\pm$0.49} \\
\ w/ exp & 20.83\tiny{$\pm$0.74} & 7.49\tiny{$\pm$1.01} & 4.42\tiny{$\pm$0.39} & 1.94\tiny{$\pm$0.54} & 1.23\tiny{$\pm$0.37} & 7.18\tiny{$\pm$0.43} & 18.85\tiny{$\pm$0.33} & 81.64\tiny{$\pm$1.87} \\
\hline
GPT-LG-medium & 27.72\tiny{$\pm$0.75} & 20.66\tiny{$\pm$0.32} & 15.47\tiny{$\pm$0.37} & 10.43\tiny{$\pm$0.66} & 6.35\tiny{$\pm$0.49} & 16.13\tiny{$\pm$0.18} & 18.77\tiny{$\pm$0.35} & 73.63\tiny{$\pm$1.23} \\
\ w/ exp & 31.11\tiny{$\pm$0.07} & 24.81\tiny{$\pm$1.62} & 17.75\tiny{$\pm$1.20} & 12.17\tiny{$\pm$0.83} & 7.55\tiny{$\pm$1.38} & 18.68\tiny{$\pm$0.94} & 25.19\tiny{$\pm$0.33} & 92.60\tiny{$\pm$1.31} \\
\hline
GPT-LG-large & 33.02\tiny{$\pm$0.93} & 29.15\tiny{$\pm$0.70} & 24.84\tiny{$\pm$1.01} & 19.15\tiny{$\pm$0.82} & 13.30\tiny{$\pm$1.20} & 23.89\tiny{$\pm$0.52} & 26.48\tiny{$\pm$0.56} & 93.80\tiny{$\pm$1.31}   \\
\ w/o pre-training & 1.99\tiny{$\pm$0.24} & 1.05\tiny{$\pm$0.98} & 0.60\tiny{$\pm$0.54} & 0.23\tiny{$\pm$0.29} & 0.17\tiny{$\pm$0.21} & 0.81\tiny{$\pm$0.32} & 1.28\tiny{$\pm$0.20} & 65.21\tiny{$\pm$15.00}  \\
\ w/o LG & 29.60\tiny{$\pm$0.36} & - & - & - & - & - & - & 95.30\tiny{$\pm$0.43} \\
\ w/ exp & 36.13\tiny{$\pm$1.26} & 32.59\tiny{$\pm$0.86} & 27.95\tiny{$\pm$3.23} & 21.42\tiny{$\pm$3.92} & 14.07\tiny{$\pm$2.87} & 26.43\tiny{$\pm$1.94} & 29.63\tiny{$\pm$0.63} & 96.96\tiny{$\pm$1.23}   \\
\ w/ ut & \textbf{42.59}\tiny{$\pm$0.94} & 37.64\tiny{$\pm$0.87} & \textbf{34.33}\tiny{$\pm$1.32} & \textbf{28.69}\tiny{$\pm$0.38} & \textbf{23.16}\tiny{$\pm$1.90} & 33.28\tiny{$\pm$0.25} & \textbf{33.40}\tiny{$\pm$0.33} & 98.49\tiny{$\pm$0.23} \\
\ w/ ut \& exp & \textbf{42.05}\tiny{$\pm$0.89} & \textbf{41.51}\tiny{$\pm$0.81} & \textbf{35.21}\tiny{$\pm$0.92} & \textbf{27.66}\tiny{$\pm$1.50} & \textbf{22.05}\tiny{$\pm$0.57} & \textbf{33.70}\tiny{$\pm$0.29} & \textbf{33.54}\tiny{$\pm$0.14} & 98.88\tiny{$\pm$0.20} \\
\bottomrule
\end{tabular}
\vspace{-2mm}
\caption{Results for CCGen.  ``exp'': explanation, ``LG'': List Generation, ``ut'': unordered training.}
\label{table:exp1}
\vspace{-3mm}
\end{center}
\end{table*}

\begin{table*}[tp!]
\scriptsize
\begin{center}
\begin{tabular}{l|ccccc|c|c|c}
\toprule
Model / Score (\%) & \textbf{1} & \textbf{2} & \textbf{3} & \textbf{4} & \textbf{5} & \textbf{Overall} & \textbf{nDCG} & \textbf{VR} \\
\midrule 
GPT-LG & 33.02\tiny{$\pm$0.93} & 29.15\tiny{$\pm$0.70} & 24.84\tiny{$\pm$1.01} & 19.15\tiny{$\pm$0.82} & 13.30\tiny{$\pm$1.20} & 23.89\tiny{$\pm$0.52} & 26.48\tiny{$\pm$0.56} & 93.80\tiny{$\pm$1.31}  \\
\ w/ exp$_{gpt2-large}$ & 35.33\tiny{$\pm$0.83} & 31.42\tiny{$\pm$1.42} & 24.44\tiny{$\pm$2.91} & 19.46\tiny{$\pm$3.86} & 12.74\tiny{$\pm$2.48} & 24.68\tiny{$\pm$1.62} & 28.90\tiny{$\pm$0.89} & 98.03\tiny{$\pm$0.40}  \\
\ w/ exp$_{opt-30b}$ & \textbf{36.13}\tiny{$\pm$1.26} & 32.59\tiny{$\pm$0.86} & \textbf{27.95}\tiny{$\pm$3.23} & 21.42\tiny{$\pm$3.92} & 14.07\tiny{$\pm$2.87} & 26.43\tiny{$\pm$1.94} & \textbf{29.63}\tiny{$\pm$0.63} & 96.96\tiny{$\pm$1.23} \\
\ w/ exp$_{opt-66b}$ & \textbf{36.81}\tiny{$\pm$1.08} & \textbf{34.59}\tiny{$\pm$0.69} & \textbf{28.52}\tiny{$\pm$1.05} & \textbf{24.44}\tiny{$\pm$0.62} & \textbf{19.97}\tiny{$\pm$1.81} & \textbf{28.87}\tiny{$\pm$0.46} & \textbf{29.26}\tiny{$\pm$0.42} & 97.96\tiny{$\pm$0.19} \\
\bottomrule
\end{tabular}
\vspace{-2mm}
\caption{Results for CCGen, with explanations distilled from different teacher models.}
\vspace{-4mm}
\label{table:diff_exp}
\end{center}
\end{table*}

\begin{table}[tp!]
    \small
    \begin{center}
    \begin{tabular}{c|ccccc|c}
        \toprule
        & \textbf{1} &
        \textbf{2} & \textbf{3} & \textbf{4} & \textbf{5} & \textbf{Overall}  \\
        \midrule
        Score & 3.02 & 2.90 & 2.85 & 2.70 & 2.63 & 2.82 \\
        \bottomrule
    \end{tabular}
    \end{center}
    \vspace{-3mm}
    \caption{Human evaluation results.}
    \label{table:human_eval}
    \vspace{-5mm}
\end{table}

{\flushleft \textbf{Implementation Details}.} See Appendix~\ref{sec:implementation}.

\subsection{Concept Generation Results}

{\flushleft \textbf{Main Results}.}
Table \ref{table:exp1} presents the results for complementary concept generation, where \textbf{1}, $\dots$, \textbf{5} denote $ACC@k_1$, $\dots$, $ACC@k_5$, \textbf{Overall} denotes $ACC@k_{overall_5}$ ($k=10$ by default), and \textbf{VR} is the \% of valid predictions since the model may generate a concept that is not in the prespecified concept set.

From the results, we find that as the scale of the base model increases (\textit{small} $\to$ \textit{medium} $\to$ \textit{large}), the performance improves significantly. In addition, $ACC@10_m$ decreases as $m$ decreases. This is reasonable since the model should have the highest confidence in its first generation.
By adding the unordered training step (w/ ut), the performance improves by a large margin, which validates the effectiveness of the two-step training strategy.

Moreover, we observe that incorporating explanations distilled from teacher models in the training process (w/ exp) not only makes the system more explainable but also improves the performance.
We think this is because a) the base model can incorporate knowledge in the explanations distilled from teacher models in its training; b) the explanations can serve as ``constraints'' to encourage the model to generate concepts that it can explain.

{\flushleft\textbf{Ablations}.}
We find that if we train the model on the same dataset from scratch (w/o pre-training), the model cannot make meaningful predictions even if it can achieve a reasonable VR. This indicates knowledge stored in the parameters of pre-trained language models is critical for CCGen.  

Furthermore, we verify the effectiveness of list generation by separating each train example to $k$ examples (w/o LG), i.e., ``\texttt{$x$ are purchased with 1) $y_1$ 2) $y_2$ ... $k$) $y_k$}'' $\to$ ``\texttt{$x$ are purchased with $y_1$}''; ``\texttt{$x$ are purchased with $y_2$}'', $\dots$, ``\texttt{$x$ are purchased with $y_k$}''. From Table~\ref{table:exp1}, we find that the $ACC@k_1$ of ``\textbf{GPT-LG-large w/o LG}'' is lower than that of ``\textbf{GPT-LG-large}''. This means that rather than 
training the model to generate a single concept each time, it is better to let the model generate a ranked list directly.

{\flushleft \textbf{Impact of Teacher Models}.} From Table \ref{table:diff_exp}, we observe that as the scale of the teacher model increases, the base model can achieve better performance in CCGen. This is because larger models tend to generate higher-quality explanations, so they can better ``teach'' the base model to explain.
Interestingly, we find that using GPT2-large itself as the teacher model can bring some performance improvement, as explained in b).

{\flushleft \textbf{Effect of Frequency}.} We analyze effect of concept frequency in Appendix~\ref{sec:exp_freq}.
The results suggest that our model performs well on low-frequent concepts.

{\flushleft \textbf{Human Evaluation}.} We hire three human annotators and conduct human evaluation on the generated concepts according the following rating scale: 
1.~unreasonable generation; 2.~reasonable generation with wrong explanation; 3.~reasonable generation with partially correct explanation; 4.~reasonable generation with reasonable explanation.
The average score (1-4) of ``\textbf{GPT-LG w/ ut \& exp}'' on 100 explanations (20 test examples) is summarized in Table \ref{table:human_eval}, with an average pairwise Cohen's $\kappa$ of 0.61 (good agreement).
This indicates the model can explain its predictions to some extent, but many explanations are still unsatisfactory, e.g., contain some factual errors -- how to improve the quality of explanations will be interesting future work.

{\flushleft \textbf{Summary}.} Our best model ``\textbf{GPT-LG-large w/ ut \& exp}'' outperforms the baselines by a large margin: not only makes much better predictions but also explains the predictions.
And the ablation study (performance of different variants of the model)  validates the effectiveness of each module.

\subsection{Sequential Generation Results}

\begin{table}[tp!]
\scriptsize
\begin{center}
\setlength\tabcolsep{4pt} %
\begin{tabular}{l|ccccc|c}
\toprule
Model / Score (\%) & \textbf{1} & \textbf{2} & \textbf{3} & \textbf{4} & \textbf{5} & \textbf{6}* \\
\midrule 
GPT-LG & 33.02 & 29.15 & 24.84 & 19.15 & 13.30 & - \\
\hline
\ + 1 & - & 41.25 & 33.62 & 25.64 & 16.95 & - \\
\ + 2 & - & - & 35.64 & 30.40 & 19.46 & - \\
\ + 3 & - & - & - & 30.20 & 23.59 & - \\
\ + 4 & - & - & - & - & 23.73 & - \\
\ + 5 & - & - & - & - & - & 17.49 \\
\hline
\ + 1 (top 10) & - & 41.65 & 34.19 & 26.35 & 17.35 & - \\
\ + 2 (top 10) & - & - & 38.69 & 32.22 & 24.33 & - \\
\ + 3 (top 10) & - & - & - & 36.35 & 28.75 & - \\
\ + 4 (top 10) & - & - & - & - & 32.56 & - \\
\ + 5 (top 10) & - & - & - & - & - & 25.75 \\
\hline
\ + 1 (all) & - & 18.12 & 20.77 & 14.47 & 13.99 & - \\
\ + 2 (all) & - & - & 20.23 & 20.09 & 16.91 & - \\
\ + 3 (all) & - & - & - & 15.73 & 17.46 & - \\
\ + 4 (all) & - & - & - & - & 18.43 & - \\
\ + 5 (all) & - & - & - & - & - & 15.44 \\
\bottomrule
\toprule
GPT-LG w/ ut & 42.59 & 37.64 & 34.33 & 28.69 & 23.16 & - \\
\hline
\ + 1 & - & 39.97 & 35.36 & 28.89 & 22.39 & - \\
\ + 2 & - & - & 38.18 & 30.63 & 22.88 & - \\
\ + 3 & - & - & - & 32.56 & 22.76 & - \\
\ + 4 & - & - & - & - & 26.32 & - \\
\ + 5 & - & - & - & - & - & 20.20 \\
\hline
\ + 1 (top 10) & - & 44.33 & 39.09 & 31.54 & 26.47 & - \\
\ + 2 (top 10) & - & - & 44.36 & 34.67 & 26.75 & -
\\
\ + 3 (top 10) & - & - & - & 39.80 & 28.80 & - \\
\ + 4 (top 10) & - & - & - & - & 35.87 & - \\
\ + 5 (top 10) & - & - & - & - & - & 33.85 \\
\hline
\ + 1 (all) & - & 42.79 & 35.70 & 31.77 & 24.87 & - \\
\ + 2 (all) & - & - & 42.39 & 35.47 & 31.05 & - \\
\ + 3 (all) & - & - & - & 41.37 & 34.96 & - \\
\ + 4 (all) & - & - & - & - & 38.89 & - \\
\ + 5 (all) & - & - & - & - & - & 40.68 \\
\bottomrule
\end{tabular}
\vspace{-2mm}
\caption{Results for sequential CCGen.}
\label{table:dynamic}
\end{center}
\vspace{-6mm}
\end{table}

In this section, we study our models in a sequential generation scenario. 
For instance, a user may browse ``\textit{Batteries}'' after browsing ``\textit{Digital Cameras}''. In this case, the model can generate complementary concepts based on both concepts with prompt ``\texttt{[SOS] Digital Cameras are purchased with 1) Batteries} 2)''. ``\textit{Battery Chargers}'' are then likely to be generated. 

Table \ref{table:dynamic} reports the results, where ``+ n'' means that the top $n$ concepts are given as input, i.e., ``\texttt{[SOS] $x$ are purchased with 1) $y_1$ 2) $y_2$ ... $k$) $y_k$ $k+1$)}'',
``+ n (top 10)'' means that we randomly sample $n$ concepts from top 10 concepts as input, and ``+ n (all)'' denotes that we randomly sample $n$ concepts from all the concepts.

For GPT-LG without unordered training (ut),
the $ACC@10_{2, \dots, 5}$ improves significantly when the top $n$ ranked concepts (or $n$ relevant concepts sampled from top 10) are given as input. This suggests that we can dynamically adjust the input of the model with more prior knowledge to achieve better performance. 
We find that if the given concept is a noisy one, the performance will decrease a lot (+ n (all)). 
However, if we adopt two-step training (\textit{GPT-LG w/ ut}), the performance is almost unaffected by the negative concepts.
This indicates that the unordered training step can make the model more robust -- 
although the previous $n-1$ concepts (either predicted or given) are wrong, the model can still make accurate prediction for the $n$th concept. 
Here we also find that for ``\textit{GPT-LG w/ ut}'', the scores of ``+ n (top 10)'' and ``+ n (all)'' are higher than those of ``+ n''. This is because for ``+ n'', the most confident predictions of the models are occupied by the given concepts. To avoid repetition, the models generate concepts that are less confident than the given ones, so the scores are lower.

{\flushleft \textit{\textbf{The $6$th Concept?!}}.} 
We provide an interesting study: without seeing any ``$6$th concept'' in training, can the model predict the $6$th concept?
We verify this by querying the model with prompt ``\texttt{[SOS] $x$ are purchased with 1) $y_1$ 2) $y_2$ ... $5$) $y_5$ $6$)}''. 
From the last column in Table \ref{table:dynamic}, we find that the models can indeed predict the $6$th concept. And the results are consistent with the results of other concepts. This means that even if the models are only trained with ranked lists of size 5, they can still predict the subsequent concepts meaningfully.

\begin{table*}[tp!]
\centering
\small
\scalebox{0.81}{
\begin{tabular}{l|p{\linewidth}}
\toprule
{\textbf{Input}} & \textbf{Camera Cases} (a concept in the test set) \\
\hline
\textit{Top 10} & 1) Batteries 2) Digital Cameras 3) Camera Lenses 4) Memory Cards 5) Lens Caps 6) Film 7) Battery Chargers 8) Screen Protector Foils 9) Camera Mounts \& Clamps    10) Dry Boxes\\
\hline
GloVe & 1) Cameras 2) Intraoral Cameras 3) On-Dash Cameras 4) Professional Video Cameras 5) Digital Voice Recorders \\
\hline
KNN & 1) Batteries 2) Cases 3) Business Card Holders 4) Rifle Cases 5) Gun Cleaning Kits \\
\hline
{GPT-LG} & 1) Tripod Heads 2) Shoe Mounts 3) Batteries 4) Camera Lenses 5) Complete Tripods  \\
\hline
{GPT-LG w/ exp} & 1) Tripod Heads: Tripod Heads are purchased with Camera Cases because they are used together. The Tripod Head is used to hold the camera and the Camera Case is used to protect the camera. 2) Complete Tripods: The Complete Tripod is a complete tripod system that includes a tripod head, a tripod foot, and a tripod collar. The tripod head is designed to be used with the camera body. The tripod foot is designed to be used with the camera body. The tripod collar is designed to be used with the tripod head. 3) Camera Mounts \& Clamps: Camera Mounts \& Clamps are used to mount the camera to the case. 4) Complete Tripods: The Complete Tripod is a complete tripod system that includes a tripod head, a tripod foot, and a tripod collar. The tripod head is designed to be used with the camera body. The tripod foot is designed to be used with the camera body. The tripod collar is designed to be used with the tripod head. 5) Batteries: Batteries are purchased with Camera Cases because the batteries are used to power the camera. The camera case is used to protect the camera from damage.  \\
\hline
{GPT-LG w/ ut} & 1) Camera Lenses 2) Camera Mounts \& Clamps 3) Batteries 4) Complete Tripods 5) Neutral Density Filters \\
\hline
{GPT-LG w/ ut \& exp} & 1) Camera Mounts \& Clamps: Camera Mounts \& Clamps are used to secure the camera to the tripod. The Camera Case is used to protect the camera from damage. 2) Batteries: Batteries are purchased with Camera Cases because the batteries are used to power the camera. The camera case is used to protect the camera from damage. 3) Lens Caps: Lens Caps are used to protect the lens of the camera from dust and scratches. 4) Tripod Heads: Tripod Heads are purchased with Camera Cases because they are used together. Tripod Heads are used to hold the camera steady while the camera is being used. Camera Cases are used to protect the camera from damage. 5) Complete Tripods: The Complete Tripod is the base of the tripod. The Camera Case is the top of the tripod. The Complete Tripod is used to hold the camera and the Camera Case is used to hold the camera and the tripod. \\
\bottomrule
\toprule
{\textbf{Input}} & \textbf{Hand Sanitizers} (a concept not in the concept set) \\
\hline
GPT-LG w/ ut \& exp & 1) Paper Towels: Paper Towels are used to clean up spills and to dry hands. Hand Sanitizers are used to clean up spills and to kill germs. 2) Exam Gloves: The gloves are used to protect the hands from germs and bacteria. The hand sanitizer is used to protect the hands from germs and bacteria. 3) Eye Wash Units: Eye Wash Units are used to clean the eyes of patients who are unable to do so themselves. Hand Sanitizers are used to clean the hands of patients who are unable to do so themselves. 4) Self-Stick Notes: Self-Stick Notes are used to write on the board. Hand Sanitizers are used to sanitize the board. 5) Disinfectants: Disinfectants are used to kill germs and bacteria. Hand sanitizers are used to kill germs and bacteria. \\
\hline
\end{tabular}
}
\caption{Sample results of different models.}
\label{table:example}
\vspace{-3mm}
\end{table*}

\subsection{Generation Examples}

Table~\ref{table:example}~(top) presents models' predictions for ``\textit{Camera Cases}''. From the results, we observe that \textit{GloVe} just finds some semantically related concepts that are not diverse and complementary to the given concept. \textit{KNN} successfully predicts ``\textit{Batteries}'', but fail in other concepts. The results of \textit{GPT-LG} already seem good, with only ``\textit{Shoe Mounts}'' failed. 
All the concepts generated by ``\textit{GPT-LG w/ exp}'' look good; however, the model repeatedly generates ``\textit{Complete Tripods}'', and the generated explanations contain some factual errors.
All the concepts predicted by ``\textit{GPT-LG w/ ut}'' and ``\textit{GPT-LG w/ ut \& exp}'' are complementary to the given concept. Furthermore, ``\textit{GPT-LG w/ ut \& exp}'' can generate reasonable explanations to justify the predictions (especially for the first four concepts).
We also examine ``\textit{Hand Sanitizers}'', a novel concept not in the concept set, and the model also does not see any ``\textit{sanitizer}'' in training. We observe that the model can generate relevant complementary concepts and explain its predictions to some extent.

\section{Related Work}

{\flushleft \textbf{Query Suggestion}.} Query suggestion is a technique used by search engines to suggests related search terms when a user enters a query~\citep{cao2008context,ma2010diversifying,ooi2015survey}.
The models are usually built upon click-through or session data in web search logs.
Notably, \citet{garg2019multiresolution,mustar2021study} apply transformer models trained with session data for query suggestion.
Different from existing works on query suggestion, which focus on suggesting surface/semantically relevant queries based on web search logs, we aim at generating \textit{complementary} concepts for a given concept through language models.

{\flushleft \textbf{Complementary Product Recommendation}.} 
Complementary product recommendation aims to recommend products that are often purchased together to serve a joint user interest \citep{mcauley2015inferring,wang2018path,zhang2018quality,kang2019complete,hao2020p,liu2020decoupled,xian2021ex3,liu2021item,angelovska2021siamese,yan2022personalized}.
For instance, 
\citet{hao2020p} develop P-Companion, where a transfer metric learning network is trained to predict the complementary products by taking product features (including title, item type) and user behavior data as input. 
\citet{yan2022personalized} propose personalized complementary product recommendation by modeling product relationships and user preferences with a graph attention network and a sequential behavior transformer.
Compared to complementary product recommendation, CCGen is a more fundamental problem that can be applied to various applications such as query suggestion and exploratory search.

{\flushleft \textbf{Open Relation Modeling}.}
Open relation modeling aims at generating sentences that describe relationships between entities/concepts \citep{huang2022open,huang-etal-2022-deer,huang2022ver,liu2022dimongen}. Analogously, our process of generating explanations can be perceived as crafting such relation descriptions between two concepts. However, the distinction lies in the nature of the relationships under consideration. In our scenario, the relation descriptions need to underscore the complementary relationship between two concepts, rather than describing any arbitrary relationship.

{\flushleft \textbf{List Question Answering}.}
\citet{katti2021question} study list question answering: given a question and a set of top $k$ URLs corresponding to the given question, extract an unordered/ordered list from the web pages to answer the question. They solve this problem by first extracting candidate lists from the given web pages and then training a classifier to rank the candidates. Unlike their work which aims to \textit{extract} lists, we propose to \textit{generate} ranked lists directly without relying on support documents.

\section{Conclusion}

In this work, we propose Complementary Concept Generation (CCGen) and present the first benchmark for this task.
We solve CCGen in a novel form of list generation. 
Specifically,
we train language models to generate ranked lists of concepts with a novel two-step training strategy.
To enable the models to explain their predictions, we incorporate explanations distilled from large teacher models.
Experimental results show that the trained models can generate high-quality concepts complementary to the input concept while producing explanations to justify the predictions.

There are various applications of our work. First, our method can be applied to query suggestion~\citep{ooi2015survey} and exploratory search~\citep{marchionini2006exploratory}, with similar formulations and objectives. The trained model can be used in (e-commerce) search engines to suggest ``\textit{complementary}'' queries, e.g., suggesting ``memory card'' and ``tripod'' for query ``digital camera''.
In addition, our system can provide useful signals for relevant downstream tasks such as complementary product recommendation by connecting products with concepts. 
Furthermore, our method suggests a way to explain product recommendation by explaining the correlation of concepts. 
In summary, the novel CCGen task, the proposed methods, and interesting findings have great potential to provide new insights for future works on these related tasks.

\section*{Limitations}

There are two main limitations of this work. 
First, the models are trained on a dataset built upon a manually crafted concept set. This ensures that most generated concepts are known and thus can be linked to associated items. However, it also limits the novelty of the generation, i.e., concepts that are not in the concept set are rarely generated. A possible solution is to automatically mine more concepts to enlarge the concept set.
Second, while explanation distillation can improve concept prediction and make the results more explainable, the generated explanations may contain factual errors, which can make them less convincing in some cases. Improving the factual correctness of the generated explanation is an interesting and important future research direction.

\bibliography{anthology,custom}
\bibliographystyle{acl_natbib}

\clearpage

\appendix

\section{Details of Dataset Construction}
\label{sec:data_detail}

Since products purchased at the same time are likely to be complements, we build a dataset for CCGen based on product co-purchase history data.
Specifically,
From the Amazon Review Data\footnote{\url{https://nijianmo.github.io/amazon/}}, we acquire the product co-purchase records with the ``\textit{also\_buy}'' attribute of products.
In Amazon Review Data, products are also associated with hierarchy structures in the ``\textit{category}'' attribute, e.g., [Electronics, Camera \& Photo, Digital Cameras, Point \& Shoot Digital Cameras] for product B00004TJ7O.
We extract concepts from these hierarchies.
In order to keep the selected concepts from being too general, e.g., ``\textit{Electronics}'', or too specific, e.g., ``\textit{Point \& Shoot Digital Cameras}'', we manually craft a prespecified set of concepts.
Specifically, we collect concepts in the categories and filter out ones with number of tokens $>6$.
Each concept is then judged by the authors on whether it is too general and too specific based on the semantic meaning and average level in the hierarchies.
To get the concept for a specific product, we traverse up from the bottom of the hierarchy to find the most fine-grained concept in the concept set. 
For example, for product B00004TJ7O, the selected concept is ``\textit{Digital Cameras}''. The size of the manually crafted concept set is $7084$.

To get the complementary concepts for each concept, we measure the likelihood of co-purchase by \textit{confidence} of co-purchase.
Specifically,
for each concept, we collect the top 10 ranked concepts based on $conf(x,y) = freq(x,y)/freq(x)$, where $freq(x,y)$ is the frequency that $x$ and $y$ are co-purchased and $freq(x)$ is the frequency of $x$.
For instance, the top 10 ranked concepts for ``\textit{Digital Cameras}'' are [Camera Lenses, Batteries, Camera Cases, Memory Cards, Battery Chargers, Lens Caps, Screen Protector Foils, Lens Hoods, Battery Grips, Remote Controls]. 
We filter out concepts with a very low frequency and randomly split concepts to build train/dev/test sets.
The statistics of the data are summarized in Table~\ref{table:data}.

\section{Implementation Details}
\label{sec:implementation}

We choose GPT-2 \cite{radford2019language} as the base model for our experiments. We test on models with different sizes, i.e., \textit{small}, \textit{medium}, \textit{large} (\textit{large} as default). 
We fix the size of ranked lists as 5 (for easier evaluation and presentation; in practice, we can set a threshold according to $conf(x,y)$ to decide the size of each ranked list).
To generate permutations for the unordered training, we randomly sample 10 permutations for each training example.
For explanation distillation, we apply Meta OPT model \citep{zhang2022opt} as the teacher model and distill explanations from the teacher model. Without additional notation, we use the 30B version\footnote{\url{https://huggingface.co/facebook/opt-30b}} (since it is can be loaded in a single 80GB A100 GPU or two 40GB A100 GPUs). To reduce the effect of hyperparameters, we apply the default hyperparameter setting with batch-size of 56, and use beam search (beam $=5$ with early stopping) in the decoding process of generation. 
For all the models and settings, we train the models with enough epochs to ensure the training converges and select the checkpoint with the best validation performance according to $nDCG_5$.
All the experiments are run with NVIDIA A100 GPUs using the HuggingFace library  \citep{wolf2020transformers}.

\section{Effect of Frequency}
\label{sec:exp_freq}

\begin{figure}[h]
  \centering
  \includegraphics[width=\linewidth]{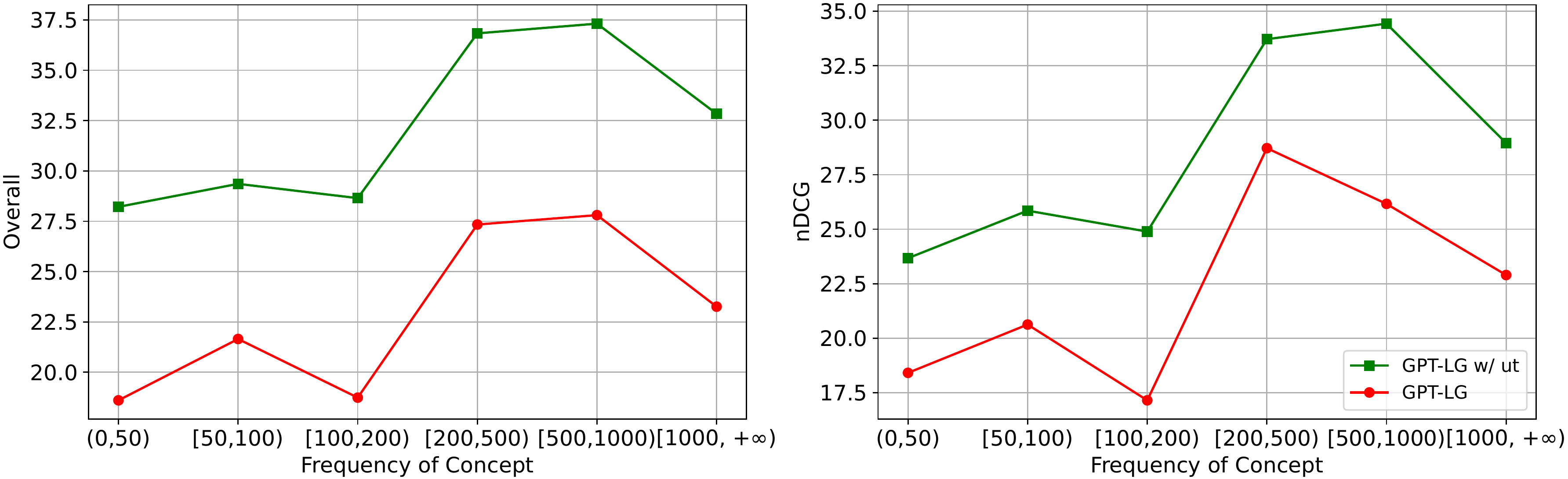}
  \caption{Results of CCGen grouped by frequency of concept. Best viewed in color.}
    \label{fig:freq_analysis}
\end{figure}

Figure~\ref{fig:freq_analysis} presents the performance of the models with respect to concept frequency. 
From the results, we observe that the results of low-frequent concepts are somewhat lower than those of high-frequent concepts. 
However, the performance is still very impressive. 
For instance, ``\textit{GPT-LG w/ ut}'' can reach an overall accuracy of $\sim$28\% for concepts with frequency $<50$.
This indicates that the proposed model can deal with ``long-tail'' items.

\end{document}